\documentclass[11pt,a4paper]{article}
\usepackage[hyperref]{ranlp2023}
\usepackage{times}
\usepackage{latexsym}
\usepackage{graphicx}
\usepackage{multirow}
\usepackage{makecell}
\usepackage{hhline}

\usepackage{hyperref}
\hypersetup{
colorlinks   = true,
citecolor    = blue,
urlcolor    =  blue
}
\usepackage{url}

\usepackage{microtype}
\usepackage{algorithm2e}
\usepackage{listings}
\usepackage{xcolor}
\usepackage{tcolorbox}
\usepackage{ragged2e}
\usepackage{tabularx}
\newcolumntype{Y}{>{\centering\arraybackslash}X}
\newcolumntype{Z}{>{\raggedleft\arraybackslash}X}
\usepackage{caption}
\usepackage{subcaption}

\lstdefinelanguage{json}{
    basicstyle=\footnotesize\ttfamily,
    commentstyle=\color{gray},
    stringstyle=\color{orange},
    morestring=[b]",
    morecomment=[l]{//},
    morecomment=[s]{/*}{*/},
    moredelim=[is][\color{blue}]{\#}{\#}
}

\usepackage{footnote}
\usepackage{tablefootnote}
\makesavenoteenv{tabular}

\newcommand{\footURL}[1]{\footnote{\url{#1}}}

\makeatletter
\newcommand\footnoteref[1]{\protected@xdef\@thefnmark{\ref{#1}}\@footnotemark}
\makeatother

\aclfinalcopy 

\title{Comparative Analysis of Named Entity Recognition\\in the Dungeons and Dragons Domain}

\author{Gayashan Weerasundara \and Nisansa de Silva \\
  Department of Computer Science \& Engineering \\
  University of Moratuwa, Sri Lanka \\
  \texttt{\{gayashan.22, NisansaDdS\}@cse.mrt.ac.lk} \\}

\date{}

\begin{document}
\maketitle
\UseRawInputEncoding
\begin{abstract}
Many NLP tasks, although well-resolved for general English, face challenges in specific domains like fantasy literature. This is evident in Named Entity Recognition (NER), which detects and categorizes entities in text. We analyzed 10 NER models on 7 Dungeons and Dragons (D\&D) adventure books to assess domain-specific performance. Using open-source Large Language Models, we annotated named entities in these books and evaluated each model's precision. Our findings indicate that, without modifications, Flair, Trankit, and Spacy outperform others in identifying named entities in the D\&D context.
\end{abstract}

\section{Introduction}
Named Entity Recognition (NER) targets the identification and classification of textual entities, such as names and locations. In the diverse and intricate vocabulary of fantasy literature, like that of Dungeons and Dragons (D\&D), NER becomes challenging~\cite{zagal2018definitions}. D\&D, a prominent fantasy literature domain, spans content for its namesake tabletop game~\cite{peiris2022synthesis,peiris2023SHADE,zhou2022ai}. These narratives inhabit fictional realms like Forgotten Realms and Dragonlance, bursting with characters, locations, and objects~\cite{gygax1974dungeons}.

NER's utility in fantasy literature is vast: from extracting information and summarizing text to character analysis and plot creation. However, conventional NER models, primarily trained on standard datasets like CoNLL-2003~\cite{tjong-kim-sang-de-meulder-2003-introduction} or OntoNotes 5.0~\cite{weischedel2013ontonotes}, might falter on fantasy texts due to their unique linguistic attributes. Recognizing the need for domain-specific adaptation, other specialized areas such as law~\citep{sugathadasa2017synergistic}, medicine~\citep{de2017discovering}, and the dynamic landscape of social media~\citep{de-silva-dou-2021-semantic} have already seen research emphasizing it. Large models, as \citet{yao2021adapt} points out, can face domain adaptation challenges, stressing the need for evaluating NER models specifically on fantasy content.

Fantasy NER has potential, especially with advancements in image generation. A notable application might involve an image generation model leveraging NER tags to derive prompts and subsequently produce contextually relevant images.

Our study contrasts 10 NER models across seven D\&D books, each averaging 118,000 words. Manual annotations of entities were made and juxtaposed against model outputs. Through precision assessments and named entity distribution analyses, we glean insights into model performances in the fantasy domain. Our key contributions include:

\begin{itemize}
\item A pioneering, comprehensive NER model evaluation on fantasy content.
\item An annotated D\&D book dataset for NER studies.
\item A deep dive into varied NER models' strengths and pitfalls in the fantasy realm.
\item Discussions on NER's role and prospects in fantasy literature.
\end{itemize}

Following this, Section 2 delves into related NER and fantasy literature works. Section 3 details our data and annotation process, while Section 4 unveils our methods and findings. Sections 5 and 6 respectively discuss insights and conclude our research, and Section 7 outlines potential future endeavors.

\section{Related Works}

NER has seen the development of various models like rule-based systems, statistical models, neural networks, and transformer-based models~\cite{seo2021controlling,liu2022chinese,krasnov2021transformer}. Although they've been trained on standard datasets, these don't encompass the complexities found in domains like fantasy literature, which poses challenges due to invented names, variable spellings, entity ambiguity, and limited resources.

We introduced a novel annotated dataset of D\&D books for NER and evaluated 10 NER models, including XLM-RoBERTa~\cite{conneau2019unsupervised}, StanfordDeID~\cite{chambon2023automated}, ELECTRA~\cite{clark2020electra}, and others.

Other studies have compared the performance of NER models on different types of texts and languages. For example, \citet{wang2021enhanced} compared Spacy, Flair, m-BERT, and camemBERT on anonymizing French commercial legal cases. They found that camemBERT performed the best overall, followed by Flair and m-BERT. SpaCy had the lowest scores but also the fastest prediction time. \cite{benesty2019ner} compared spaCy, Flair, and Stanford Core NLP on anonymizing English court decisions. They found that Flair had the highest scores, followed by Stanford Core NLP and spaCy. \cite{shelar2020named} compared rule-based, CRF-based, and BERT-based techniques for NER on text data. They found that BERT-based technique had the highest accuracy and recall, followed by CRF-based and rule-based techniques. \cite{naseer2021named} compared NLTK, spaCy, Stanford Core NLP, and BERT~\cite{devlin2018bert} on extracting information from resumes. They found that BERT had the highest accuracy and F-measure, followed by spaCy, Stanford Core NLP, and NLTK.

These studies suggest that different NER models may have different strengths and weaknesses depending on the type, language, and domain of the text data. Our study aims to contribute to this understanding by providing the first systematic comparison of NER models on fantasy texts and analyzing their performance and characteristics.

\section{Data Collection and Annotation}

This section details the data sources and annotation process utilized for our named entity recognition (NER) task, a subtask of information extraction that classifies named entities in unstructured text into categories such as persons, organizations, and locations~\cite{mohit2014named}.

We examined seven adventure books from the Dungeons and Dragons (D\&D) realm, listed in table~\ref{tab:book list}. These books, primarily adventure-centric, were sourced from the official DnDBeyond site, the main publication hub for D\&D by Wizards of the Coast.

Through a comprehensive analysis of these texts, we used their rich narratives and character dynamics to benchmark and assess various NER models in this intricate domain.




\begin{table}[h!]
\resizebox{\columnwidth}{!}{%
\begin{tabular}{lrr}
\hline
\textbf{Book}  & 
\multicolumn{2}{c}{\textbf{Counts}} \\
\hhline{~--}
& \textbf{Words} & \textbf{Topics} \\
\hline

Lost mine of Phandelver~\cite{baker2014lost}     & 45947               & 29                     \\
Hoard of the Dragon Queen~\cite{baur2014hoard} & 74243               & 45                     \\
Rise of Tiamat~\cite{baur2014rise}                & 80065               & 48      \\ 
Curse of Strahd~\cite{perkins2016curse}            & 154519              & 62                     \\
Tomb of Annihilation~\cite{Perkins2017Tomb}       & 148605              & 35                     \\
Candlekeep Mysteries~\cite{Perkins2021candlekeep}          & 141104              & 106                    \\
The Wild Beyond the Witchlight~\cite{allan2021wild}  & 184135              & 60                     \\
\hline     
\end{tabular}%
}
\caption{D\&D adventure books}
\label{tab:book list}
\end{table}

Each of our chosen books averages 118,000 words. The selection was driven by our familiarity with these tales and the broader D\&D universe. Additionally, they span multiple genres, themes, and settings in the fantasy realm, offering a vast array of named entities for NER.

The source books were transformed into text and organized hierarchically into chapters, topics, and paragraphs. An example from "The Wild Beyond the Witchlight" is displayed in table~\ref{tab:my-table}.



\begin{table}[!htb]
\resizebox{\columnwidth}{!}{%
\begin{tabular}{llll}
\hline
\textbf{Chapter} &
  \textbf{Topic} &
  \textbf{Paragraph} &
  \textbf{Word Count} \\
\hline
\multirow{7}{*}{\begin{tabular}[c]{@{}l@{}}Introduction:\\ Into the\\Feywild\end{tabular}} &
  \multirow{5}{*}{\begin{tabular}[c]{@{}l@{}}Adventure \\ Summary\end{tabular}} &
  \begin{tabular}[c]{@{}l@{}}The main \\antagonists of \\this story are \\ three hags...\end{tabular} &
  131 \\
 &
   &
  \begin{tabular}[c]{@{}l@{}}One of the \\many novelties\\ of this adventure \\is that...\end{tabular} &
  43 \\
 &
   &
  \begin{tabular}[c]{@{}l@{}}The characters \\are drawn into \\ the adventure \\by one of two \\ adventure hooks. \\You choose...\end{tabular} &
  31 \\
 &
   &
  \begin{tabular}[c]{@{}l@{}}Chapter 1 \\describes the \\ Witchlight Carnival...\end{tabular} &
  40 \\
 &
   &
  ... &
  ... \\
 &
  \multirow{2}{*}{\begin{tabular}[c]{@{}l@{}}Running the \\ Adventure\end{tabular}} &
  \begin{tabular}[c]{@{}l@{}}The Monster \\Manual contains\\ stat blocks \\for most of the\\ creatures encountered \\in this...\end{tabular} &
  72 \\
 &
   &
  \begin{tabular}[c]{@{}l@{}}Spells and \\equipment mentioned \\ in the adventure \\are described \\ in the Player’s \\Handbook...\end{tabular} &
  31
\end{tabular}%
}
\caption{Content hierarchy in a book}
\label{tab:my-table}
\end{table}

We first manually perused the source books, marking named entities hierarchically by chapter, topic, and paragraph, recording only entity counts. Subsequently, we employed three state-of-the-art large language models: Bloom~\cite{scao2022bloom}, OpenLLaMA~\cite{openlm2023openllama}, and Dolly~\cite{databricks2023dolly}, to detect named entities in each book chapter. These models, trained on vast conversational data, can craft natural language responses, making them apt for the intricate language patterns in D\&D texts, such as neologisms and metaphors.

After eliminating duplicates and pinpointing unique entities, we verified these results against our initial counts. The named entities identified by the three LLMs underwent a manual review for accuracy and consistency, adding crucial missed entities. Table~\ref{tab:LLM comparison} contrasts the named entity counts from each LLM, with recall metrics based on entities common across all models.



\begin{table*}[!htb]
\begin{tabularx}{\textwidth}{|l|ZZ|ZZ|ZZ|Z|}
\hline
\textbf{Book} & \multicolumn{2}{c|}{\textbf{Bloom}} & \multicolumn{2}{c|}{\textbf{Dolly}} & \multicolumn{2}{c|}{\textbf{OpenLLaMA}} & \textbf{Total Unique} \\
\hhline{~------~}
& Count & Recall & Count & Recall 
& Count & Recall & \textbf{Entities} \\ 
\hline
Lost Mine of   Phandelver     & 21  & 0.47 & 32 & 0.73 & 40  & 0.91 & 44  \\
Hoard of the Dragon Queen     & 58  & 0.89 & 62 & 0.95 & 60 & 0.92 & 65  \\
Rise of Tiamat                & 54  & 0.88 & 57 & 0.93 & 53 & 0.87 & 61  \\
Curse Of Strahd               & 92  & 0.90 & 96 & 0.94 & 101 & 0.99 & 102 \\
Tomb of Annihilation          & 101 & 0.80 & 99 & 0.79 & 112 & 0.89 & 126\\
Candle keep Mysteries         & 60  & 0.87 & 61 & 0.88 & 64  & 0.93 & 69  \\
The Wild Beyond Witch   Light & 66  & 0.84 & 67 & 0.85 & 71  & 0.89 & 79  \\
\hline
\end{tabularx}
\caption{Result comparison between LLMs}
\label{tab:LLM comparison}
\end{table*}

When annotating the resultant named entities we followed a set of annotation guidelines that define the entity types and the annotation rules for our NER task. The entity types that were used are:

\begin{itemize}
\item Person: any named character or creature that can act as an agent, such as heroes, villains, allies, enemies, etc.
\item Organization: any named group or faction that has a common goal or identity, such as guilds, cults, clans, etc.
\item Location: any named place or region that has a geographical or spatial dimension, such as cities, dungeons, forests, etc.
\item Misc: any named entity that do not belong to above mentioned categories. (This contain important information like Spells, Artifacts, Potions etc.)
\end{itemize}

The process of annotation is done through a script, where a paragraph segment is taken iteratively and fed into the LLMs with a template prompt.

\begin{table}[!htb]
\resizebox{\columnwidth}{!}{%
\begin{tabular}{|lll|}
\hline
\multicolumn{3}{|l|}{\begin{tabular}[c]{@{}l@{}}Please identify and list all named entities \\ in the following text using the BIO \\ (beginning-inside-outside) scheme:\end{tabular}} \\ \hline
\multicolumn{3}{|l|}{\begin{tabular}[c]{@{}l@{}}"The traveling extravaganza known as \\ the Witchlight Carnival visits your \\ world once every eight years. You \\ have a dim memory of sneaking into \\ the carnival as a child without paying...\\ ...pair of elves named Mister Witch and \\ Mister Light—were decidedly unhelpful."\end{tabular}} \\ \hline
\multicolumn{1}{|l|}{B-Organization:} &
  \multicolumn{1}{l|}{Witchlight} &
  Carnival \\ \hline
\multicolumn{1}{|l|}{I-Person:} &
  \multicolumn{1}{l|}{Mister} &
  Witch \\ \hline
\multicolumn{1}{|l|}{I-Person:} &
  \multicolumn{1}{l|}{Mister} &
  Light \\ \hline
\end{tabular}%
}
\caption{Process of Annotation}
\label{tab:anotation prompt}
\end{table}

Following Algorithm~\ref{alg:NER} is the pseudo-code for the process in identifying named entities:

\begin{algorithm}[!htb]
\textbf{Input:} Books\;
\textbf{Output:} Named entities\;
\ForEach{book}{
  segments $\gets$ divideIntoSegments(book)\;
  \ForEach{segment in segments}{
    paragraphs $\gets$ divideIntoParagraphs(segment)\;
    \ForEach{paragraph in paragraphs}{
      \ForEach{LLM in LLMs}{
        prompt $\gets$ createPrompt(paragraph)\;
        namedEntities $\gets$ LLM(prompt)\;
        processNamedEntities(namedEntities)\;
      }
    }
  }
}
removeDuplicates(namedEntities)\;
\caption{Named Entity Recognition using Multiple LLMs}
\label{alg:NER}
\end{algorithm}

As shown in above pseudocode, the algorithm~\ref{alg:NER} takes a set of books as input and outputs the named entities identified by the LLMs. The algorithm iterates over each book and divides it into segments. Each segment is further divided into paragraphs, and each paragraph is iteratively fed into each of the LLMs with a prompt to identify named entities. The named entities identified by each LLM are then processed and saved. Finally, all named entities are checked for duplicates, and those duplicates are removed.

After named entities were recognized, they were then mapped in to json objects for storage as shown in Figure~\ref{fig:json}. Nesting of objects is done according to the hierarchy as mentioned in table~\ref{tab:my-table}. Each of the named entities were nested in an array of entities as entity objects with corresponding attributes as mentioned bellow.

\begin{figure}[!htb]
\begin{tcolorbox}[colback=white,colframe=black!75!black,title=JSON object]
\begin{lstlisting}[language=json]
{
    "book": "Candlekeep Mysteries",
    "chapter": 1,
    "text": "The Book of Inner Alchemy is one of Candlekeep’s...",
    "entities": [
        {
            ...
        },
        {
            "entity": "B-Location",
            "score": 0.9659823,
            "index": 8,
            "word": "Candlekeep",
            "start": 42,
            "end": 51
        },
    ]
}
\end{lstlisting}
\end{tcolorbox}
\caption{sample format of the JSON output}
\label{fig:json}
\end{figure}

\section{Experimental Setup and Results}

The experiment was conducted to identify how effective are the NER models when using them as off the shelf models in identifying named entities for a fantasy domain when there are no available corpora for fine tuning. For testing we used 10 different contemporary NER midels.

Following table~\ref{tab:statistics} shows the identified count of named entities for each categories of the adventure book Candlekeep Mysteries.

\begin{table}[!htb]
\resizebox{\columnwidth}{!}{%
\begin{tabular}{llllll}
\hline
\textbf{Model} & \textbf{PER} & \textbf{LOC} & \textbf{ORG} & \textbf{MSC} & \textbf{All} \\
\hline
XLM-RoBERTa~\cite{conneau2019unsupervised}                 & 16 & 0  & 3 & 4  & 23 \\
StanfordAIMI~\cite{chambon2023automated}  & 0  & 0  & 1 & 18 & 19 \\
ELECTRA~\cite{clark2020electra}                     & 10 & 0  & 1 & 10 & 21 \\
WikiNEuRal~\cite{tedeschi2021wikineural} & 23 & 4  & 6 & 1  & 34 \\
BERT~\cite{devlin2018bert}                        & 9  & 1  & 1 & 0  & 11 \\
RoBERTaNER~\cite{baptiste2022roberta}   & 1  & 0  & 0 & 17 & 18 \\
BERT-CRF~\cite{souza2019portuguese}                    & 12 & 0  & 0 & 0  & 12 \\
Flair~\cite{akbik2018coling}                       & 28 & 14 & 6 & 4  & 54 \\
Spacy~\cite{honnibal2017spacy}                       & 21 & 11 & 7 & 18 & 57 \\
Trankit~\cite{nguyen2021trankit}                     & 25 & 15 & 2 & 2  & 44 \\
\hline
\end{tabular}%
}
\caption{Statistics for the adventure book Candlekeep Mysteries. The NER tags are as follows, Person: PER, Location: LOC, Organization: ORG, and Miscellaneous: MSC}
\label{tab:statistics}
\end{table}

The testing approach for the NER models mirrors algorithm~\ref{alg:NER}. Here, paragraphs of input text are fed into the models without specific prompts. The resultant output is refined by filtering out corrupted values (e.g., "Strahd Von Zarovich" might be mistakenly split into two distinct names) and redundant entries, before being transitioned into the JSON structure showcased in Figure~\ref{fig:json}.

During initial processing, NER models often produce numerous erroneous outputs. These arise from factors like incomplete word detection, missegmentation of terms, or misinterpretation of special characters. Such discrepancies can be mitigated using string manipulations and by cross-referencing outputs with a pre-curated list of named entities.

Figure~\ref{fig:json currupted} displays entries that encountered corruption. These highlight instances where NER models incorrectly processed and extracted entities from the source material.




\begin{figure}[!htb]
\begin{tcolorbox}[colback=white,colframe=black!75!black,title=JSON object]
\begin{lstlisting}[language=json]
{
    "entities": [
        {
            "entity": "I-Person",
            "score": 0.5659823,
            "index": 308,
            "word": "Fembris L#",
            "start": 1160,
            "end": 1162
        },
        {
            "entity": "I-Person",
            "score": 0.51227564,
            "index": 309,
            "word": "rlancer",
            "start": 1162,
            "end": 1164
        }
    ]
}
\end{lstlisting}
\end{tcolorbox}
\caption{sample format of a corrupted JSON outputs}
\label{fig:json currupted}
\end{figure}

In the given example shown in Figure~\ref{fig:json currupted}, the name "Fembris Larlancer" is erroneously divided into two distinct words, "Fembris L\#" and "rlancer", as a result of corruption during the NER processing stage. This example underscores the challenges faced during the entity extraction process and the need for robust post-processing to ensure the accuracy and quality of the extracted entities.

After removing corrupted and eligible named entities, duplicate entries must be removed to do a proper comparison of performance between different models. For this tuples of words in adjacent positions were generated and compared. For example Mayor Lei Duvezin, Mayor Duvezin, Lei Duvezin and Duvezin all refers to the same entity with the label Person. In cases such as above tuple with most similarity matches will be retained as the named entity and duplicates will be removed.

To visualize the raw named entity identification potential of each model, a density plot was plotted with respect to count of identified named entities with NER models. Following Figure~\ref{fig:density plot} shows the density of named entities recognized by each NER model. The hue represents the overlapping count ranges of named entities identified in each source book.

\begin{figure}[!htb]
\centering
\includegraphics[width=0.45\textwidth]{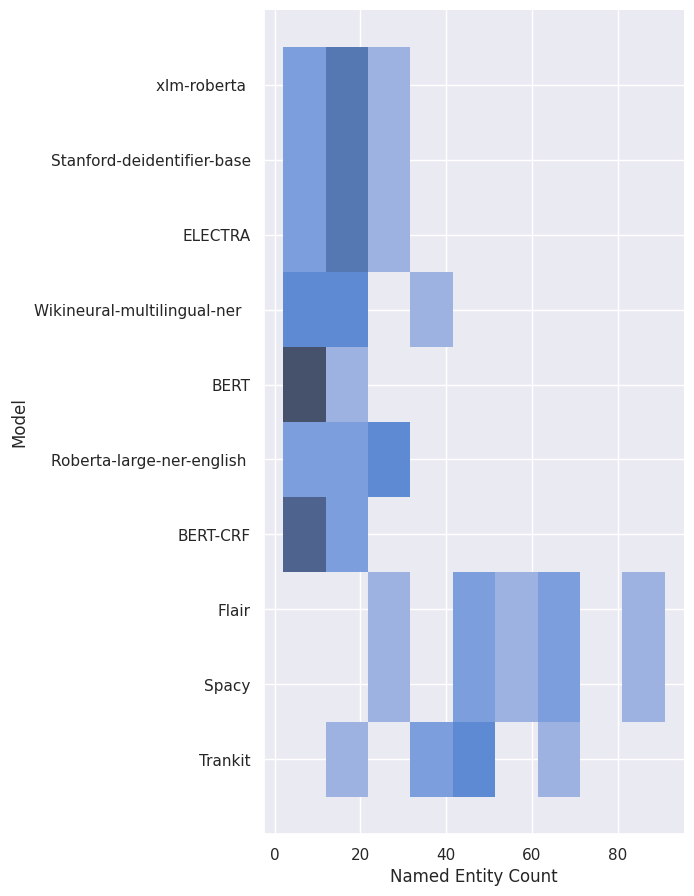}
\caption{Density plot for each model}
\label{fig:density plot}
\end{figure}

Without training, NLP frameworks like Trankit~\cite{nguyen2021trankit}, Flair~\cite{akbik2018coling}, and Spacy~\cite{honnibal2017spacy} show a strong baseline in entity recognition.

Model precision is key in performance evaluation. This is gauged by comparing the true positive entities with actual named entities. This comparison can be visually represented for each model across source books.

For a comprehensive model assessment across books, Kernel Density Estimation (KDE) is used. It's a non-parametric method estimating the probability density function~\cite{terrell1992variable}:




\[f(x) = \frac{1}{nh} \sum_{i=1}^n K\left(\frac{x - x_i}{h}\right)\]

where:

\begin{itemize}
\item xi are the data points
\item K is the kernel function, which is typically a Gaussian function or a uniform function
\item h is the bandwidth, which determines the width of the kernel function and controls the smoothness of the estimate
\item n is the number of data points
\end{itemize}

KDE calculates $f(x)$ through a summed kernel function $K(u)$, anchored at data points $x_i$.

\begin{figure*}[!htb]
\centering
\includegraphics[width=0.95\textwidth]{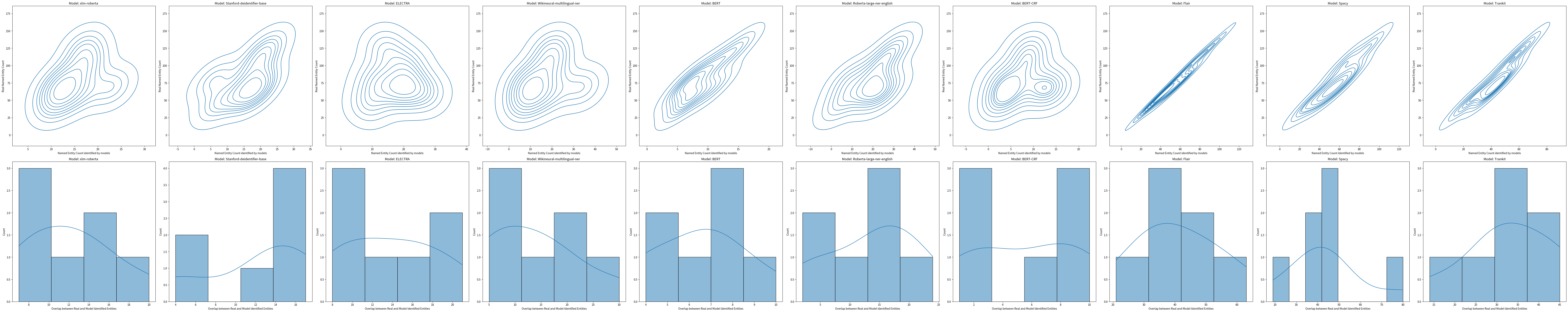}
\caption{Distribution plot for each model}
\label{fig:distribution plot}
\end{figure*}

Figure~\ref{fig:distribution plot} illustrates models' efficacy over seven source books. A gradient near 1 signifies optimal performance.

\begin{figure*}[!htb]
\centering
 \begin{subfigure}[t]{0.44\textwidth}
    \includegraphics[width=\textwidth]{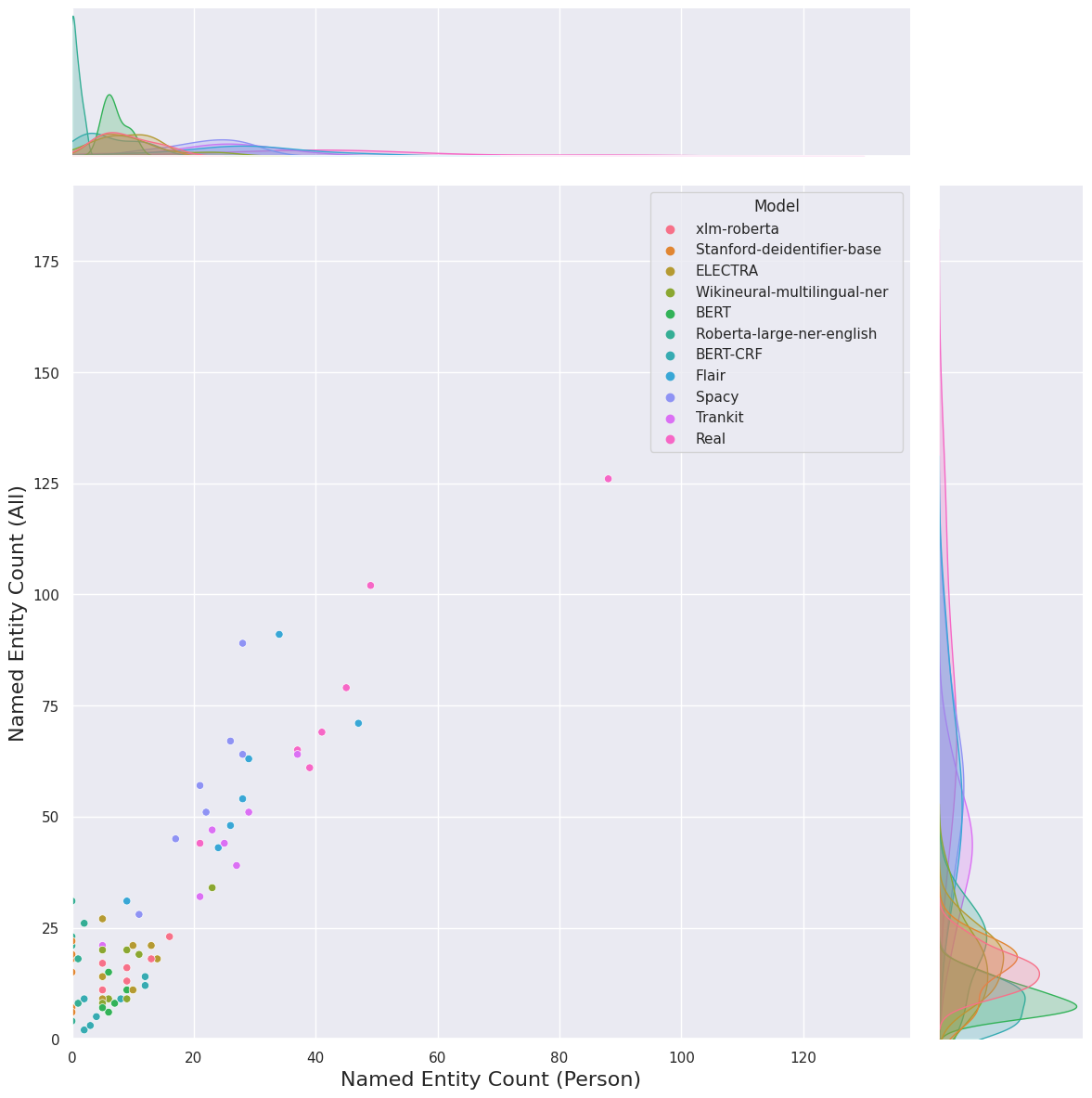}
    \caption{Models}
    \label{fig:freqModels}
 \end{subfigure}%
 \hfill
 \begin{subfigure}[t]{0.44\textwidth}
    \includegraphics[width=\textwidth]{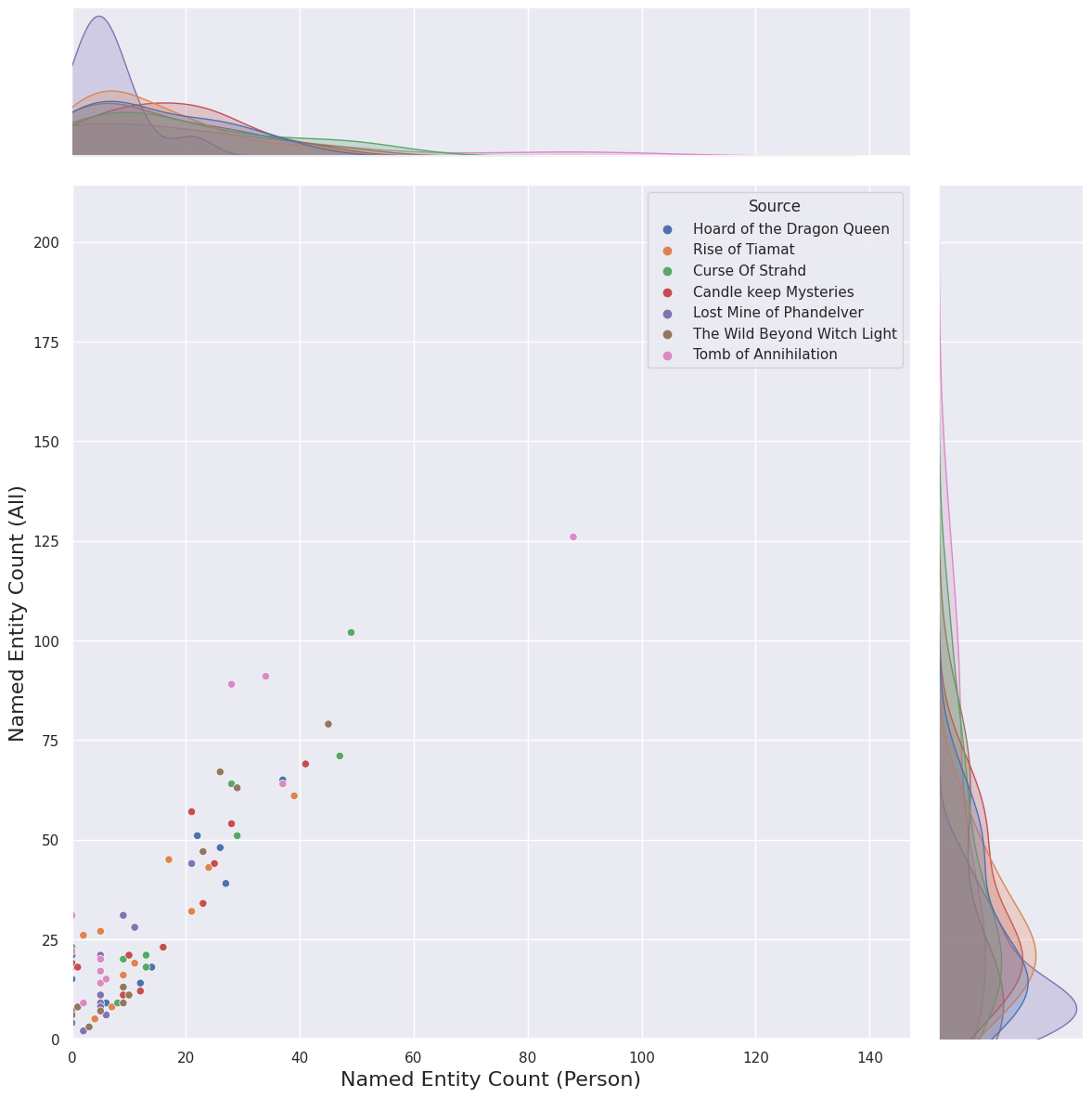}
    \caption{Adventure sourcebooks.}
    \label{fig:freqSbooks}
  \end{subfigure}%
  \caption{Frequency plots with respect to models and adventure sourcebooks}
  \label{fig:frequencyPlots}
\end{figure*}

In D\&D, named characters, with their elaborate backstories, are central. Assessing a model's inclination to identify these characters over other entities is vital. This inclination can be visualized by juxtaposing character counts with total entities, contrasted against real metrics. Figure~\ref{fig:frequencyPlots} delineates the frequency of character identification across all source books. Meanwhile, Figure~\ref{fig:freqModels} and Figure~\ref{fig:freqSbooks} depict the distribution pertaining to models and books, respectively.

In the D\&D landscape, named characters, renowned for their intricate histories, are paramount. Evaluating a model's propensity to spot these characters in relation to other entities is imperative due to the significant role characters play in D\&D narratives. This bias can be graphically represented by mapping character counts against all identified entities and contrasting them with authentic counts. By scrutinizing the named entity counts from diverse NER models and comparing them to true values, one can infer model behavior and efficacy. Figure~\ref{fig:frequencyPlots} offers a glimpse into character recognition frequency for different models across sourcebooks, with Figure~\ref{fig:freqModels} and Figure~\ref{fig:freqSbooks} charting the distributions for models and books respectively.



From Figure~\ref{fig:freqSbooks}, we observe a consistent ratio between characters and other named entities across books. This consistency allows us to downplay book variability and focus on the insights from Figure~\ref{fig:freqModels}. Notably, NLP frameworks such as Spacy~\cite{honnibal2017spacy} and Flair~\cite{akbik2018coling} exhibit more balanced frequency distributions, indicating a higher character identification ratio. Although this might be unfavorable in certain contexts, in this domain, aligning character identifications closely with overall named entity values signals optimal performance. This suggests Spacy and Flair perform exceptionally in an off-the-shelf setting.

\begin{figure}[!htb]
\centering
\includegraphics[width=0.45\textwidth]{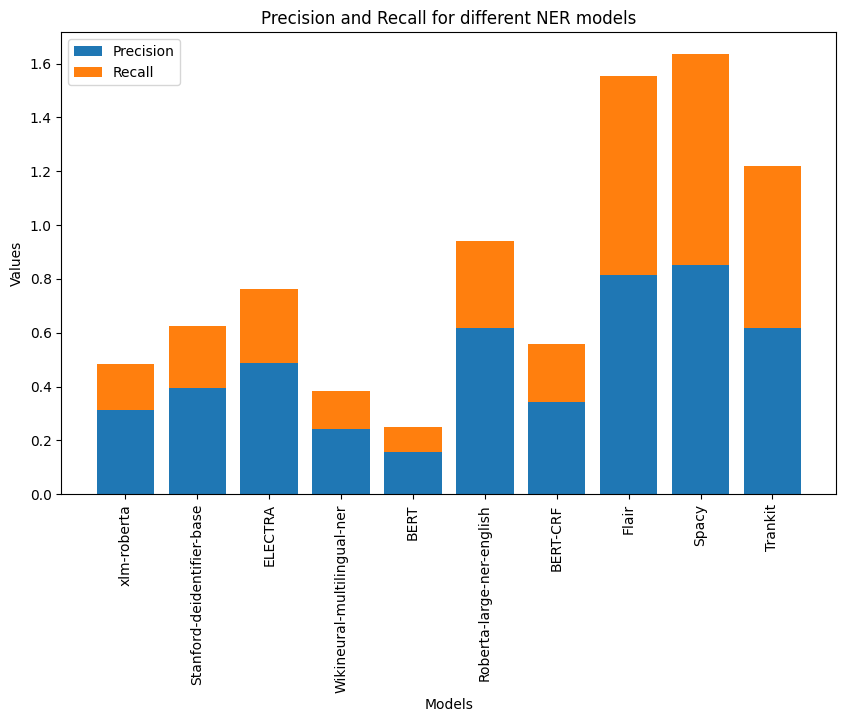}
\caption{Precision graph for different NER models}
\label{fig:precision graph}
\end{figure}

Figure~\ref{fig:precision graph} showcases precision and recall metrics for each NER model. To determine recall, we derived the true positive count from average unique named entity counts, while the true count originated from LLM models, as outlined in table~\ref{tab:LLM comparison}. For precision, false positives were ascertained from misidentified unique named entities on average.

The precision and recall values were averaged for each model across source books, and plotted to offer a concise visualization of each NER model's performance.

Evidently, Flair and Spacy outshine other NER models in precision and performance, while Trankit~\cite{nguyen2021trankit} excels in recall relative to its precision.

\section{Discussion}

We undertook a Named Entity Recognition (NER) task on seven adventure books from the esteemed Dungeons and Dragons (D\&D) series. Our methodology involved manual entity annotations in these books, which were subsequently verified against outputs from three leading language models: Bloom, OpenLLaMA, and Dolly.

Our annotation guidelines delineated entity types into categories like person, organization, location, and misc. Ten NER models were subsequently employed to gauge their efficacy in recognizing named entities within D\&D. Among these, Flair, Trankit, and Spacy emerged superior, mirroring findings from past NER-centric studies. Conversely, StanfordDeID~\cite{chambon2023automated} and RoBERTaNER~\cite{baptiste2022roberta} lagged in performance. A precision-centric analysis further reiterated the dominance of Flair, Trankit, and Spacy over their counterparts.

The findings imply that while generic models can decently handle NER tasks in specialized domains like D\&D, performance inconsistencies exist across models. Employing annotation guidelines bolsters consistency in entity recognition across varied books and contexts. Moreover, incorporating large language models for automated annotations can significantly mitigate the manual intervention needed for comprehensive datasets, particularly in intricate domains such as D\&D.

However, our study bears certain caveats. We refrained from fine-tuning the NER models specifically for D\&D, so our findings are indicative of generic model capabilities and might not capture the full potential of domain-specific optimization. Our dataset, comprising just seven books, might not encompass the depth and breadth of D\&D narratives. The exclusive focus on Wizards of the Coast publications could also inadvertently introduce stylistic biases. Finally, while our study zeroes in on D\&D as a fantasy subset, our insights might not seamlessly extend to other literary domains with their unique nuances.

\section{Conclusion}

Our exploration illuminates the remarkable potential of harnessing off-the-shelf models for NER tasks within the D\&D universe's nuanced realm. Some models showcase an impressive baseline in entity recognition for this domain without extensive fine-tuning. However, there's a compelling need for continued research and refinement to tailor these models optimally for D\&D's unique intricacies.

Additionally, our research serves as a foundational resource for future inquiries. The dataset we've curated and our annotation guidelines stand as a benchmark for gauging the efficiency of future NER models or techniques. Consequently, our work not only reveals the current prowess of NER models within the D\&D context but also sets the stage for continued innovation at the confluence of fantasy literature and artificial intelligence.


\section{Future Works}

Based on our findings and limitations, we suggest some directions for future research. One direction is to fine-tune NER models on the D\&D dataset and comparing their performance with off-the-shelf models. Additionally, other techniques such as transfer learning or domain adaptation could be explored to improve the performance of NER models in the D\&D domain. Another direction is to use different data sources for NER in D\&D, such as novels, comics, podcasts, or video games. A third direction is to apply different evaluation metrics for NER in D\&D, such as F1-score, recall, accuracy, or error analysis. Finally other aspects of NER in D\&D can also be explored, such as entity linking, coreference resolution, relation extraction, or sentiment analysis.



\bibliographystyle{acl_natbib}
\bibliography{anthology,ranlp2023}


\end{document}